\documentclass[letterpaper, 10 pt, conference]{ieeeconf}  

\IEEEoverridecommandlockouts                              

\usepackage{graphics} 
\usepackage{epsfig} 
\usepackage{mathptmx} 
\usepackage{times} 
\usepackage{amsmath} 
\usepackage{amssymb}  
\usepackage{caption}
\title{\LARGE \bf
Part-Level 3D Gaussian Vehicle Generation with Joint and Hinge Axis Estimation
}

\author{Shiyao Qian$^{1}$, Yuan Ren$^{1}$, Dongfeng Bai$^{1}$, Bingbing Liu$^{1}$ 
\thanks{$^{1}$The authors are with Huawei Noah’s Ark Lab.
        {\tt\small \{shiyao.qian, yuan.ren3, baidongfeng, liu.bingbing\}@huawei.com}}%
\thanks{$^{2}$Shiyao Qian is with the University of Toronto and contributed to this work during an internship at Huawei Noah’s Ark Lab, Canada.
        {\tt\small shiyao.qian@mail.utoronto.ca}}%
 }

\begin{document}

\maketitle
\thispagestyle{empty}
\pagestyle{empty}

\begin{abstract}
Simulation is essential for autonomous driving, yet current frameworks often model vehicles as rigid assets and fail to capture part-level articulation. With perception algorithms increasingly leveraging dynamics such as wheel steering or door opening, realistic simulation requires animatable vehicle representations. Existing CAD-based pipelines are limited by library coverage and fixed templates, preventing faithful reconstruction of in-the-wild instances.

We propose a generative framework that, from a single image or sparse multi-view input, synthesizes an animatable 3D Gaussian vehicle. Our method addresses two challenges: (i) large 3D asset generators are optimized for static quality but not articulation, leading to distortions at part boundaries when animated; and (ii) segmentation alone cannot provide the kinematic parameters required for motion. To overcome this, we introduce a part-edge refinement module that enforces exclusive Gaussian ownership and a kinematic reasoning head that predicts joint positions and hinge axes of movable parts. Together, these components enable faithful part-aware simulation, bridging the gap between static generation and animatable vehicle models.
\end{abstract}

\section{INTRODUCTION}

Over the past few years, simulation has played a crucial role in autonomous driving, particularly in supporting the training and validation of perception algorithms. Recent advances such as 3D Gaussian splatting (3DGS) and neural rendering have made simulation both photorealistic and efficient, enabling large-scale synthetic data generation for diverse driving scenarios. Early perception methods primarily focused on predicting the 3D bounding box of an entire vehicle, without the capability to detect the operation of individual components. More recently, however, perception algorithms have advanced to capture part-level dynamics, such as wheel steering or door opening, which can be leveraged to infer vehicle trajectories or anticipate passenger actions. These emerging capabilities place higher demands on simulation frameworks, which must now not only model whole vehicles but also faithfully capture the articulation and motion of individual parts.

Prior work such as UrbanCAD simulates articulated vehicles by inserting CAD assets into the scene. However, these pipelines are fundamentally constrained by the coverage of the CAD library: the inserted vehicle often deviates from the instance in the source video, and the articulation patterns are entirely dictated by the chosen CAD template. To address this limitation, we propose a generative method that, given a single image or sparse multi-view observations, synthesizes an animatable 3D Gaussian representation of the vehicle and infers per-part motion. While recent foundation models can generate generic 3D Gaussian assets from images (large 3D asset generation) and part-level segmentation methods such as Segment Any 3D Gaussians (SAGA) can segment those assets into semantic components, a complete pipeline for in-the-wild, animatable vehicles still requires solving two open issues. First, current large 3D asset generators optimize for overall static quality and are not articulation-aware: once parts are animated, geometry around edges and hinges often exhibits distortions such as tearing and self-intersection, which degrade realism. Second, given a part segmentation, we still need to automatically identify movable parts and infer their kinematics, including joint position and hinge axis direction. 

To address these challenges, we make two contributions. First, we propose a part-edge refinement algorithm that explicitly suppresses cross-part Gaussian sharing: Gaussians near segmentation boundaries are rotated or re-scaled to enforce exclusive part ownership. This prevents color/opacity bleeding across parts. Second, we develop a kinematic reasoning head that, given part segments, automatically identifies movable parts and predicts their joint positions and hinge axes. Together, these components close the gap from static asset generation to animatable vehicle models.

The rest of this paper is structured as follows. Section II surveys related work on 3D part-level segmentation and autonomous driving simulation. Section III provides a concise review of prior works that our framework builds upon. Section IV presents the proposed method, including part-edge refinement and motion mode estimation. Section V reports experimental results, while Section VI discusses key findings, limitations, and future directions. Finally, Section VII concludes the paper.

\section{RELATED WORK}

\subsection{3D Part-level Segmentation}
Going beyond semantic segmentation of entire scenes, part-level segmentation of 3D objects has emerged as a key research direction for understanding object structure and enabling part-level editing. Fueled by datasets such as ShapeNetPart \cite{shapenet2015} and PartNet \cite{Mo_2019_CVPR}, early works—including the PointNet series \cite{Qi_2017_CVPR, qi2017pointnet++} and other geometry-based CNNs \cite{li2018pointcnn, phan2018dgcnn, thomas2019kpconv}—focused on supervised 3D part segmentation by learning geometric features directly from point cloud data.

To overcome the limited class diversity in annotated 3D datasets, recent approaches leverage feature lifting from 2D to 3D, transferring advances from large-scale 2D vision models into low-shot and open-vocabulary 3D segmentation \cite{radford2021learningtransferablevisualmodels, Li_2022_CVPR, oquab2023dinov2, kirillov2023segment}. Among these, the Segment Anything Model (SAM) has become a popular 2D foundation model, applied across point clouds \cite{ma20253d, deng2025geosam2unleashingpowersam2, zhou2023partslipenhancinglowshot3d}, meshes \cite{tang2025segmentmesh, yang2024sampart3dsegment3dobjects}, and 3DGS representations \cite{ye2024gaussiangroupingsegmentedit, guo2024semantic, piekenbrinck2025opensplat3d, cen2025segment3dgaussians}.

Different methods extend SAM in diverse ways: SAMPart3D \cite{yang2024sampart3dsegment3dobjects} distills DINOv2 features \cite{oquab2023dinov2} into its 3D backbone and retrieves part labels via a multimodal large language model; Find3D \cite{ma20253d} trains a transformer backbone using SigCLIP embeddings \cite{zhai2023sigmoidlosslanguageimage} of Gemini-generated labels \cite{team2024gemini}, enabling queryable point-level semantic features; PartSLIP++ \cite{zhou2023partslipenhancinglowshot3d} generates bounding boxes from text-prompted CLIP \cite{li2022groundedlanguageimagepretraining} to guide SAM for controllable 2D masks; and OpenSplat3D \cite{piekenbrinck2025opensplat3d} incorporates MasQCLIP \cite{xu2023masqclip} for per-instance CLIP features. Meanwhile, GaussianGrouping \cite{ye2024gaussiangroupingsegmentedit} and SAGA \cite{cen2025segment3dgaussians} adopt a common strategy of using SAM-derived masks as ground truth to train 3D features specialized for 3DGS segmentation.

Our method builds on SAGA, further integrating PointNet++ to enable automatic part retrieval and motion axes prediction, extending 3DGS segmentation toward 3D asset part articulation.

\subsection{Autonomous Driving Simulation}
Realistic simulation of driving scenes forms the foundation for developing safe and scalable self-driving systems. To construct reliable digital twin scenes of real-world scenarios, dynamic actors such as vehicles are essential. Traditional car simulators \cite{Dosovitskiy17, Gaidon_2016_CVPR, shah2017airsim} are typically built on game engines, providing fine-grained control over scene layout and dynamic actors while maintaining scalability. However, the heavy manual effort and inherent artifacts in scene construction introduce a gap between these virtual environments and the real world, limiting their ability to faithfully reconstruct edge cases.

With advances in neural rendering \cite{Martin-Brualla_2021_CVPR, oechsle2021unisurf, niemeyer2022regnerf}, data-driven simulation methods \cite{zhang2021ners, manivasagam2020lidarsim, guo2020object} have gained traction by offering realistic and faithful simulations from sparse data input. Yet, the 3D vehicles produced by these methods exhibit limited controllability, making them unsuitable for closed-loop data generation or out-of-distribution (OOD) scenario simulation (e.g., cars with open doors).

To address this, CADSim \cite{wang2023cadsimrobustscalableinthewild} leverages a small set of part-aware CAD models to reconstruct vehicle geometry with controllable articulated wheels. However, it does not extend this controllability to other parts such as front doors. UrbanCAD \cite{lu2025urbancadhighlycontrollablephotorealistic} further highlights the importance of part editability by showing significant performance drops in self-driving perception models when shifting from normal to OOD data of the same vehicle instances. It also introduces a framework for CAD model retrieval and material optimization to construct 3D vehicle digital twins. Despite improvements in photorealistic scene insertion, UrbanCAD’s reliance on fixed CAD geometries creates inconsistencies between the inserted models and their real-world counterparts.

Our work seeks to push the boundary of the fidelity-controllability tradeoff by leveraging 3D model generation and segmentation techniques to construct vehicle digital twins that offer greater manipulability—including front doors and wheels—while preserving high consistency with the original cars.

\section{PRELIMINARY}

In this section, we briefly introduce three prior works that provide the basis for our method, including
\subsection{Large 3D Asset Generation Model (TRELLIS)}
Xiang et al. \cite{xiang2025structured3dlatentsscalable} introduced TRELLIS, a novel method for efficient and high-quality 3D asset generation. At its core is the Structured LATent (SLAT) representation, a unified latent space that can be decoded into multiple output formats, including 3D Gaussians. Trained on a large dataset of 500K collected 3D assets, TRELLIS achieves state-of-the-art performance on image-prompted generation, significantly outperforming prior approaches across standard metrics.

The generation pipeline employs a rectified flow model, which first produces a voxelized sparse structure as a coarse representation of the asset. This coarse structure is then refined into structured latents, which can be decoded into the final 3D asset. Such a two-stage design provides greater control over geometry and supports tuning-free editing: minimal changes in the prompt or sampling process yield consistent yet flexible modifications.

These properties make TRELLIS particularly well-suited to our work -- part-level manipulations demand both geometric consistency and structural awareness in the generated vehicle assets.

\subsection{Segment Any 3D Gaussians (SAGA)} \label{saga_preliminary}
SAGA \cite{cen2025segment3dgaussians} enables promptable segmentation of 3DGS models by introducing a trainable contrastive feature vector for each Gaussian. The pipeline begins by applying the SAM model to multi-view renderings of a 3DGS object, extracting 2D masks and their scales. These masks and scales then supervise a scale-aware contrastive learning stage, where sampled pixels from rendered contrastive features are trained against the corresponding mask identities. In effect, this lifts 2D mask information into the 3D feature space.

Once trained, segmentation is performed in the contrastive feature space using HDBSCAN clustering \cite{mcinnes2017hdbscan}. Importantly, the training part is fully decoupled from the earlier steps of asset preparation and mask retrieval, allowing any modifications or interpolations in those stages without disrupting the final segmentation.

SAGA also introduces a vote-based open-vocabulary extension: similar masks are clustered into 3D targets, and masked images within each cluster are encoded with CLIP features to enable query- or label-based segmentation. However, this extension has shown limited effectiveness for our specific part retrieval task, as discussed further in Sec.~\ref{PointNet_query}.

\subsection{PointNet++}
The PointNet series has served as a backbone for numerous learning-based 3D segmentation methods. PointNet \cite{Qi_2017_CVPR} pioneered segmentation on sparse, irregular point cloud data. Its core idea is to apply shared neural networks followed by a max-pooling operation to obtain an order-invariant point set feature, expressed as:
\begin{equation}
f(x_1, x_2, \ldots, x_n) = \gamma \left( \max_{i = 1, \ldots, n} { h(x_i) } \right)
\label{eq:pointnet}
\end{equation} 
where $\gamma$ and h are usually multi-layer perceptron (MLP) networks and $\{ x_1, x_2, \ldots, x_n \}$ is the unordered point set feature input.

\begin{figure*}[t]
    \centering
    \includegraphics[width=1.0\textwidth]{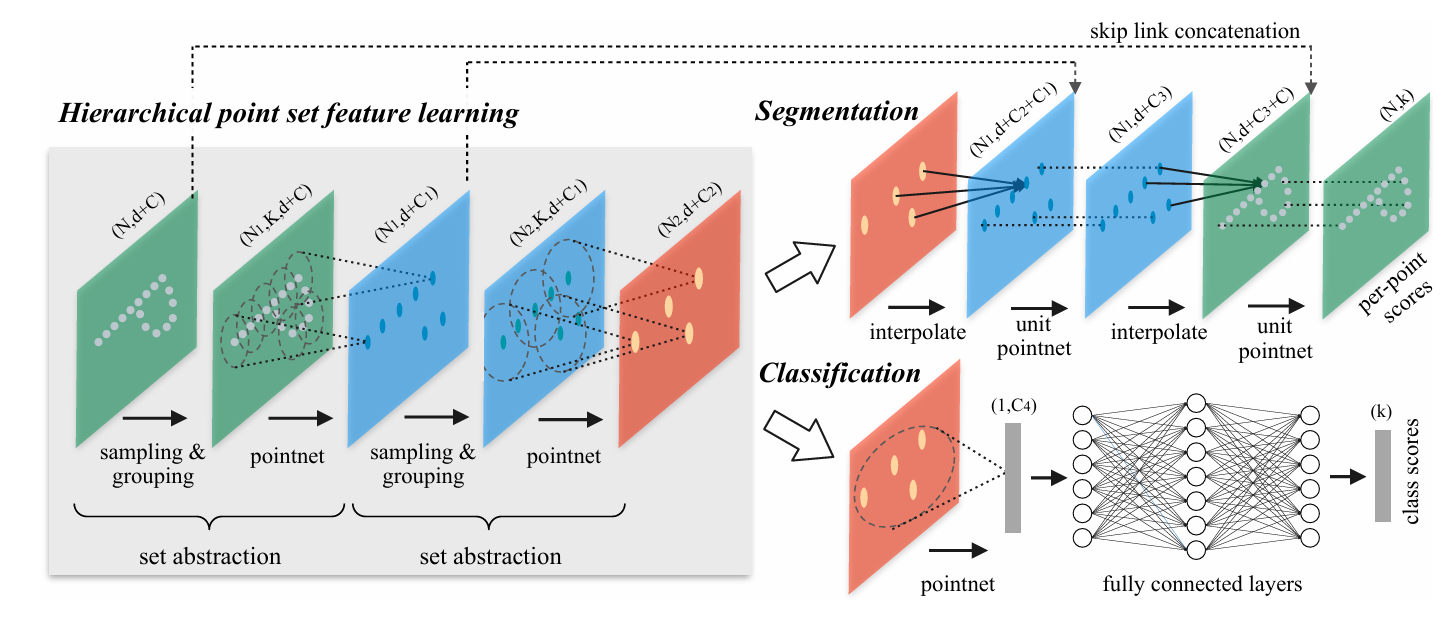}
    \caption{Overview of the PointNet++ hierarchical feature learning architecture. For segmentation, learned features are interpolated back to the original point set to produce per-point scores. For classification, an additional abstraction layer aggregates features into a global representation, which is passed through fully connected layers to predict object class scores.}
    \label{pointnet2_architecture}
\end{figure*}

PointNet++ \cite{qi2017pointnet++} extends this framework with a hierarchical structure and multi-scale grouping, enabling more robust and fine-grained learning of local geometry. This design improves shape classification and segmentation accuracy compared to the original PointNet. An overview of the architecture is provided in the original paper (Fig.~\ref{pointnet2_architecture}).

Subsequent works have sought to address limitations in PointNet++ feature learning, such as potential information loss in max-pooling when aggregating local neighborhoods. For instance, DGCNN \cite{wang2019dynamicgraphcnnlearning} and PointConv \cite{wu2020pointconvdeepconvolutionalnetworks} enhance local feature extraction by dynamically constructing graphs or modeling convolutional kernels on point sets.

Despite these advances, our work adopts PointNet++ as a lightweight and reliable backbone. Its maximum pooling and identical convolution layers across neighborhoods are well-suited to handle the potential local point displacement gap between training and inference data in our pipeline—a point we will elaborate on in Sec.~\ref{PointNet_training_data}.

\begin{figure*}[t]
    \centering
    \includegraphics[width=1.0\textwidth]{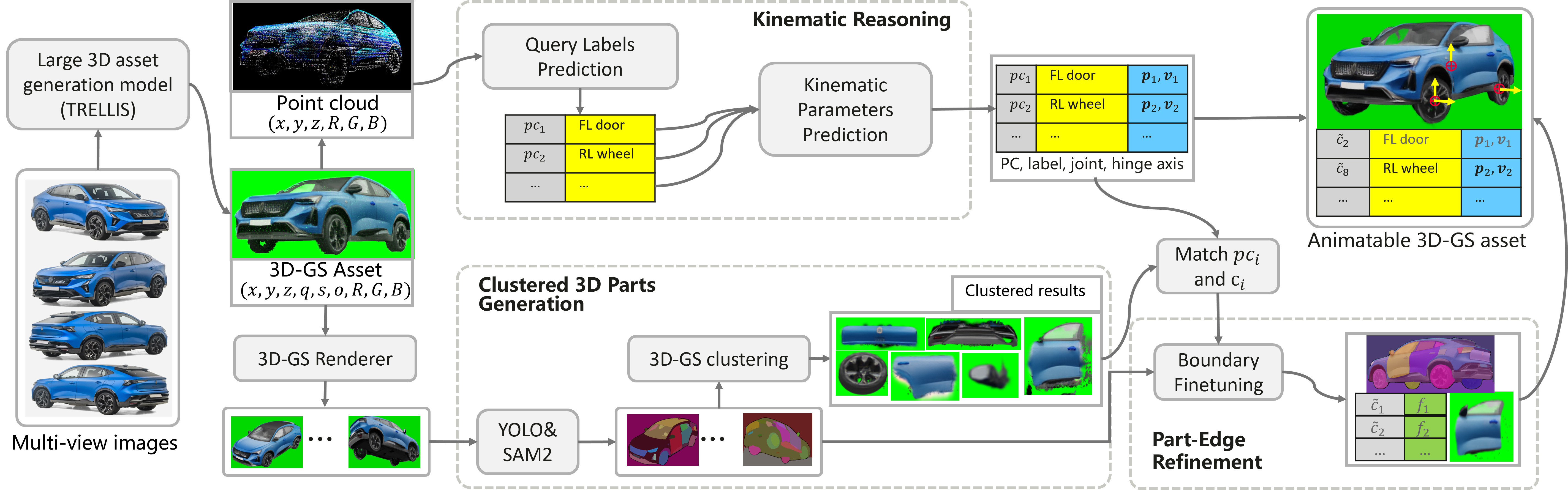}
    \caption{Overview of our pipeline. TRELLIS generates a 3DGS asset from four multi-view images of the vehicle. The asset is converted into a point cloud, which is processed by the kinematic reasoning module for part-level segmentation and motion axes estimation. In parallel, SAGA segments the 3DGS representation directly, with additional movable part masks integrated alongside auto-segmented ones. The resulting clusters are aligned with the PointNet++ segmentation outputs to retain six target parts. Finally, part geometries are refined at the boundary in the part-edge refinement module, and all segmented parts with their predicted motion axes are combined to produce an animatable 3DGS vehicle.}
    \label{overview}
\end{figure*}

\section{METHOD}
An overview of our method is shown in Fig.~\ref{overview}. We first generate a 3D Gaussian vehicle asset using a large-scale generative model. The asset is then processed in parallel by the kinematic reasoning module and the augmented SAGA pipeline. The kinematic reasoning module employs two PointNet++ networks, trained independently from scratch on our collected dataset: one performs part-level segmentation to assign cluster-query labels, while the other predicts kinematic parameters governing vehicle motion. In parallel, candidate part clusters are produced by the SAGA pipeline, enhanced with complete movable part mask insertions. By aligning the segmentation outputs with the clustering results, we assemble a 3DGS part representation in which each part is defined by a Gaussian cluster and its associated motion parameters (joint position or rotation axis). Finally, a part-edge refinement module adjusts Gaussian geometry near boundaries, enforcing alignment with movable-part masks and suppressing floating artifacts.

\subsection{Clustered 3D Parts Generation}
Our car asset segmentation pipeline builds on the SAGA framework \cite{cen2025segment3dgaussians}, but replaces its 3DGS reconstruction step with TRELLIS generation \cite{xiang2025structured3dlatentsscalable} and incorporates the more advanced 2D vision model SAM2 \cite{ravi2024sam2segmentimages}. To enable more controllable segmentation of movable car parts, we further integrate masks predicted by a pretrained You Only Look Once (YOLO) \cite{redmon2016lookonceunifiedrealtime} model and propagate them across views using SAM2.

\subsubsection{TRELLIS Asset Generation}
While 3DGS reconstruction can achieve high-fidelity assets, it typically requires a large number of input images captured from continuous viewpoints and incurs significant computational cost. Furthermore, the training process of 3DGS reconstruction \cite{kerbl20233dgaussiansplattingrealtime} primarily focuses on aligning rendered outputs with input images. With only limited views of a car, these methods often fail to preserve disentangled geometry in unseen regions—an essential property for part editing. For instance, reconstructed wheels are frequently distorted from their cylindrical form, making them unsuitable for realistic rotation or steering.

In contrast, generative models such as TRELLIS \cite{xiang2025structured3dlatentsscalable}, conditioned on only a few reference images, can complete unseen regions more plausibly while retaining fine geometric details necessary for part manipulation. At the same time, TRELLIS is significantly more computationally efficient and preserves strong consistency with the input views. For this reason, we adopt TRELLIS to generate the initial 3DGS vehicle asset, and render 400 images from uniformly sampled viewpoints on a unit sphere around the model for subsequent masking and training steps.

\subsubsection{Controlled Part Mask Insertion}
In the original SAGA pipeline, SAM is used to automatically generate masks for contrastive feature training. Although SAM’s multi-granularity segmentation is effective for retrieving scale-aware contrastive features, it lacks prior knowledge of target shapes. For example, the front door should include the door body, window, and side mirror, yet SAM often segments these into separate regions based solely on visual similarity, making it difficult to retain the complete structure. To address this, we introduce task-specific masks for target car parts, ensuring that the final clustering results align with our intended part definitions. In addition, we replace SAM with the more advanced SAM2, which provides faster inference speed and improved accuracy.

To obtain accurate masks for doors and wheels, we first apply YOLO trained on the Car Parts Dataset \cite{DSMLR_Carparts}, which contains 500 annotated images of sedans, pickups, and SUVs with 18 instance-level masks per image, including our target parts. Using this model, we generate initial mask predictions on selected views of the 3D asset. These masks are then propagated across neighboring views with sufficient visibility using SAM2 video segmentation, which supports mask tracking from YOLO prompts. For more reliable prompting and tracking, each front door is divided into two masks—one for the rear-view mirror and one for the main body.

Finally, we merge the predicted movable-part masks with the SAM2 auto-segmentation results by appending the new masks and removing overlapping originals. This ensures that SAGA receives complete part masks in most training views, enabling the network to consistently cluster each movable car part as a single component.

\subsection{Kinematic Reasoning Module}
\label{pointnet2_method}

\subsubsection{Training Data} \label{PointNet_training_data}
As 3D Gaussians are a relatively new representation for differentiable rendering, the scarcity of labeled segmentation data is particularly severe. Most existing segmentation approaches for 3DGS, including SAGA (which we adopt as a backbone), operate in a zero-shot setting and thus cannot provide labels for querying. To overcome this limitation, we turn to the point cloud representation, which benefits from wider data availability and more established learning-based segmentation methods.

Among open-source resources, 3DRealCar \cite{du20253drealcarinthewildrgbdcar} is the only dataset offering 3D car parsing annotations covering all required parts. However, its labels are derived from 2D parsing maps predicted by SAM, making them too noisy for reliable training. Moreover, annotations for joint positions and hinge axes are absent.

\begin{figure}[t]
    \centering
    \includegraphics[width=0.4\textwidth]{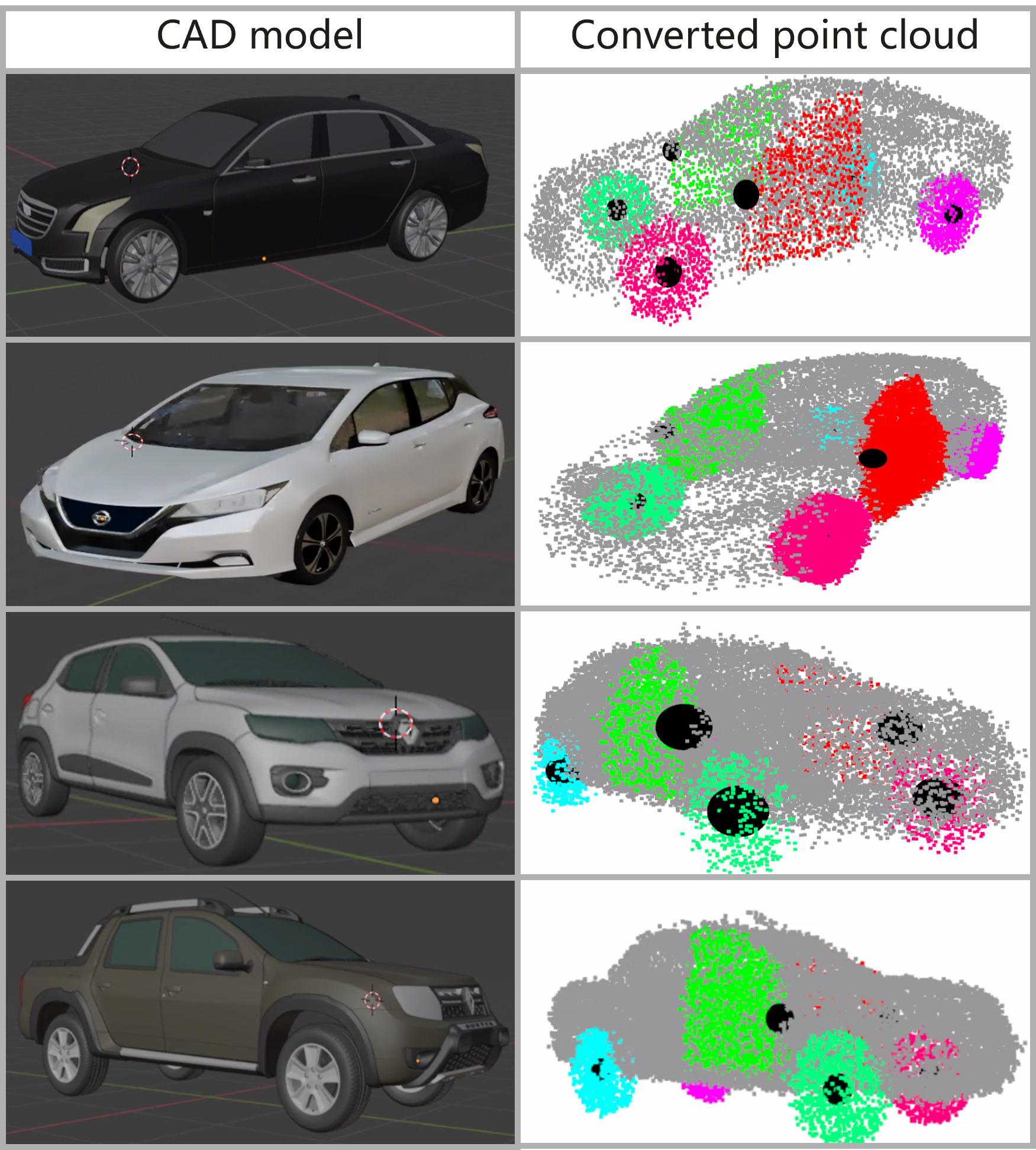}
    \caption{PointNet++ training data examples. Point cloud data are colored according to their part label, with joint and hinge positions marked as black balls.}
    \label{pointnet2_dataset}
\end{figure}

To address these gaps, we construct a custom dataset from 100 high-quality free CAD car models. We manually annotate the front door and wheel meshes, then convert them into point cloud data by sampling on the asset surfaces, while preserving part labels and ground-truth joint and hinge positions. Example samples are shown in Fig.~\ref{pointnet2_dataset}. At inference time, we similarly convert 3D Gaussians into point cloud form using their mean positions and view-independent RGB values.

\subsubsection{Query Labels Prediction} \label{PointNet_query}
Among the dozens of clusters produced by HDBSCAN in SAGA's contrastive feature space, it is crucial to identify those corresponding to the editable parts of interest. As discussed in Sec.~\ref{saga_preliminary}, although SAGA does provide an open-vocabulary segmentation extension, this approach fails to encode spatial information of car parts (e.g., distinguishing the left door from the right door), often retrieving the same cluster for different part queries. its accuracy also remains low due to the multiple intermediate steps required between a text query and the final cluster retrieval.

To overcome these limitations, we introduce a supervised part segmentation strategy to automatically query the required clusters. For robustness, we adopt PointNet++ \cite{qi2017pointnet++}, whose globally shared feature extraction mechanism across point sets helps mitigate discrepancies between training data (point clouds sampled from mesh surfaces) and inference data (point clouds sampled from 3D Gaussian centers).

Details of the complete query point prediction procedure and additional gap bridging techniques are provided in Sec.~\ref{query_label_prediction_details}

\subsubsection{Kinematic Parameters Prediction}
A key step toward enabling part editability in our 3DGS model is predicting the motion axes of wheels and front doors. For this task, we again adopt the PointNet++ architecture, but in its classification mode (see Fig.~\ref{pointnet2_architecture}), which produces a global feature vector from the input point set. We modify the final MLP layers to output hinge axes and joint positions rather than class scores. This design allows the network to directly regress the joint positions and hinge axis positions from the input vehicle point cloud.

Further details of the design are provided in Sec.~\ref{kinematic_parameters_prediction_details}.

\subsection{Part-Edge Refinement Module}
With the additional inserted masks and the automatic querying from PointNet++, we can retrieve the correct car parts from the SAGA clustering results. However, these clusters often contain blurry boundaries and floating Gaussians, which become particularly noticeable after part editing. Inspired by \cite{hu2025sagdboundaryenhancedsegment3d}, we propose a simple but novel loss function to finetune the geometry of editable parts, ensuring better alignment with the movable part mask inputs.

To produce cleaner parts, we first render the selected clusters from the same views used for masking. We then penalize the alpha channel values of pixels outside the corresponding part masks, encouraging floating Gaussians to attach back to the part and reorienting edge Gaussians so they align with boundaries instead of bleeding across them. This term is defined as:
\begin{equation}
    L_{\text{outside}} = \left( \frac{1}{H \times W} \right) \times \sum_{i,j} \left[ \alpha(i,j) \times (1 - M(i,j)) \right]
    \label{eq:L_outside}
\end{equation}
where H and W are the image dimensions, $\alpha(i, j)$ is the alpha channel value at pixel $(i, j)$, and $M(i, j)$ is corresponding 2D binary mask.

To prevent the optimization from overfitting to mask boundaries and altering the part’s overall appearance, we render the original, unoptimized part Gaussians and use their in-mask regions as ground truth. A simple L1 photometric loss is then added, defined as:
\begin{equation}
    L_{\text{photo}} = \frac{\sum_{i,j} \|R(i,j) - GT(i,j)\| \times M(i,j)}{\sum_{i,j} M(i,j)}
    \label{eq:photo_loss}
\end{equation}
where $R(i,j)$ is the rendered RGB color of the optimized part and $GT(i,j)$ is the corresponding ground-truth rendering.

During optimization, we fix Gaussian opacities and colors to avoid artifacts such as transparent boundaries or color shifts, focusing solely on geometry refinement. The final objective combines the photometric and mask-boundary terms as:
\begin{equation}
    L_{\text{optimization}} = (1 - \lambda_{\text{outside}}) L_{\text{photo}} + \lambda_{\text{outside}} L_{\text{outside}}
    \label{eq:optimization_loss}
\end{equation}
Here, $\lambda_{\text{outside}}$ controls the relative contribution of the two terms. In our geometry optimization process, we set $\lambda_{\text{outside}}=0.5$, giving equal weight to both losses.

\begin{figure}[t]
    \centering
    \includegraphics[width=0.4\textwidth]{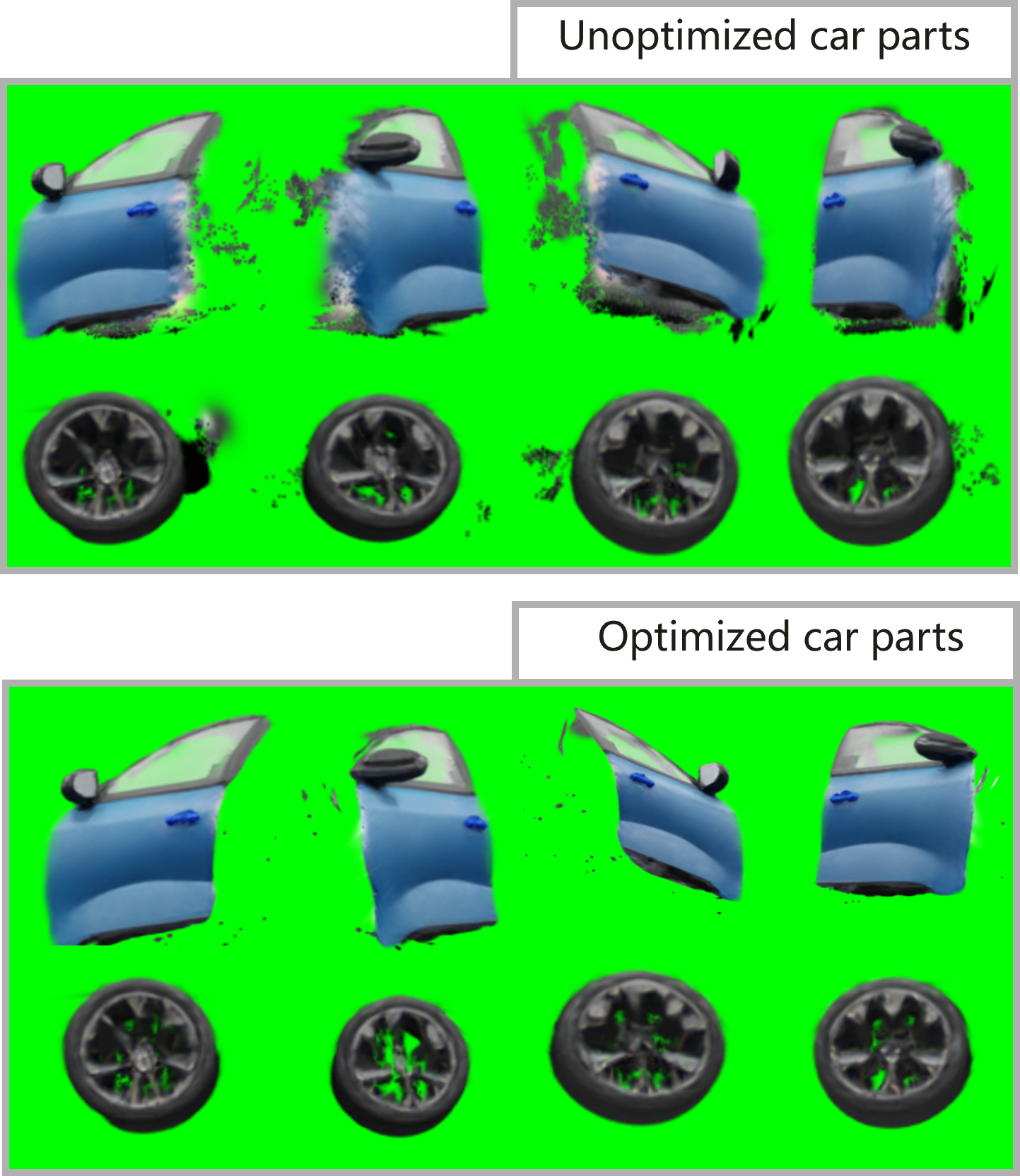}
    \caption{Rendering of all car parts before and after the boundary finetuning step}
    \label{finetune_result}
\end{figure}

This finetuning process yields cleaner and sharper part clusters that are better suited for subsequent editing. A direct comparison between parts before and after optimization is shown in Fig.~\ref{finetune_result}.

\section{EXPERIMENTS}

\subsection{Dataset}
We evaluate part editing using second-generation CARLA vehicle models \cite{Dosovitskiy17} with controllable wheels and doors. Four special vehicles outside our scope are excluded, resulting in seven usable models. For each model, we generate eight distinct part states (door opening, wheel steering, and rotation), producing a total of 56 ground-truth vehicle assets for evaluation.

Direct comparisons with prior work are not feasible: UrbanCAD shares a similar focus but does not address part segmentation and lacks publicly available code, while other works either focus on zero-shot segmentation (e.g., SAGA) or emphasize general part motion reasoning \cite{yu2024gamma, xiang2020sapien, wang2019shape2motion}. We therefore conduct an ablation study to analyze the contribution of each component in our pipeline.

\subsection{Ablation Study}
We compare pipeline variants with different components enabled. The evaluation measures similarity between generated and ground-truth vehicle renderings under matched camera poses and part states. Visual fidelity is assessed using CLIP feature similarity \cite{radford2021learningtransferablevisualmodels} and LPIPS \cite{zhang2018unreasonableeffectivenessdeepfeatures}, with results summarized in Table~\ref{tab:ablation_study}.

Incremental improvements are observed as additional modules are introduced. Mask insertion helps cluster parts more completely, though residual boundary noise and floating artifacts limit performance gains. The final boundary finetuning step produces significantly cleaner edges and reduces floating Gaussians, leading to the best quantitative performance. Although the absolute changes in CLIP and LPIPS appear small, they are meaningful in our setting, where only six segmented movable parts contribute to the gain in overall similarity.

\subsection{Qualitative Comparison}
\begin{figure*}[t]
    \centering
    \includegraphics[width=1.0\textwidth]{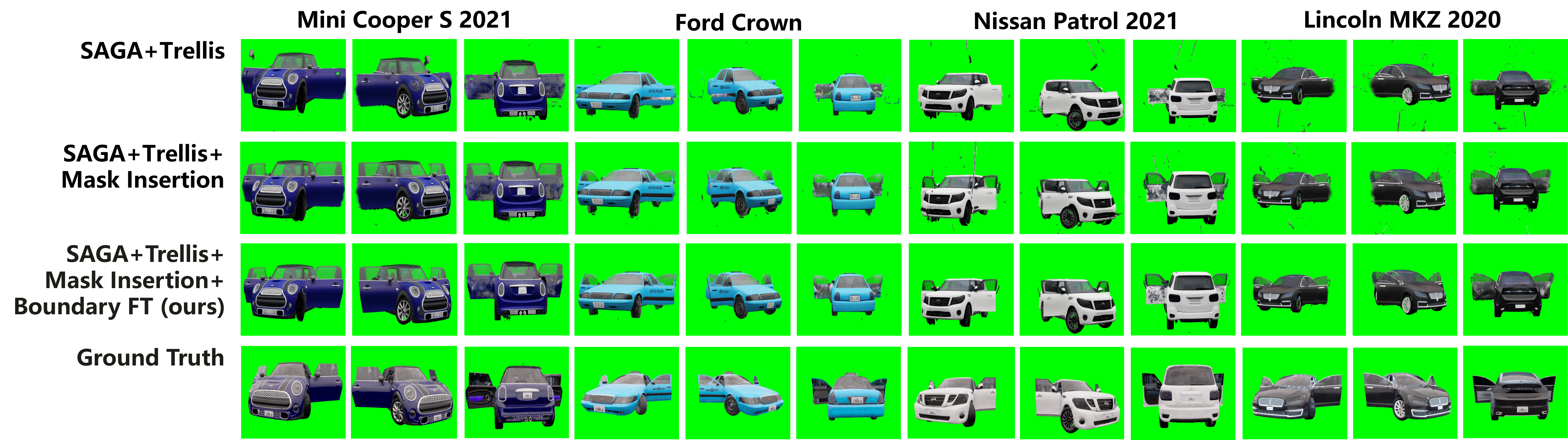}
    \caption{Qualitative comparison of part-manipulated vehicle renderings across different ablation pipelines. From top to bottom: (1) SAGA + Trellis, (2) + Mask Insertion, (3) + Boundary Finetuning (ours), and (4) Ground Truth. Mask insertion leads to more complete part clustering, while the final boundary finetuning module produces cleaner edges and reduces floating Gaussians, yielding renderings that align more closely with the ground-truth vehicles in perceptual quality.}
    \label{qualitative result}
\end{figure*}

Fig.~\ref{qualitative result} illustrates renderings from the same ablation variants. Without mask insertion, edits are often incomplete (e.g., door bodies manipulated while windows remain fixed). With mask insertion, parts are clustered more fully but boundaries remain noisy. Incorporating boundary finetuning yields clean, well-defined parts that visually align with the manipulated ground-truth vehicles, consistent with the quantitative improvements in Table~\ref{tab:ablation_study}. Further details of the experimental setup are provided in Sec.~\ref{experimental_details}.

\begin{table}[t]
    \centering
    \begin{tabular}{|p{2.3cm}|c|c|}
        \hline
        Components & CLIP$(\uparrow)$ & LPIPS$(\downarrow)$ \\
        \hline
        TRELLIS + SAGA & 0.848 & 0.305 \\
        \hline
        TRELLIS + SAGA + 
        Mask insertion & 0.855 & 0.304 \\
        \hline
        TRELLIS + SAGA + 
        Mask insertion + 
        Edge ft (ours) & 0.873 & 0.283 \\
        \hline
    \end{tabular}
    \caption{Ablation study of our pipeline. Performance with different components enabled are recorded. Adding mask insertion improves similarity, while the full pipeline with boundary finetuning produces the best result -- the highest CLIP similarity $(\uparrow)$ and lowest LPIPS distance $(\downarrow)$.}
    \label{tab:ablation_study}
\end{table}

\section{LIMITATIONS AND FUTURE WORK}
This work has several limitations that suggest concrete directions for improvement. Our part-level supervision and edge refinement are driven exclusively by 2D segmentation results. Without explicit constraints on part thickness, articulated panels (e.g. doors) can degenerate into near-zero-thickness sheets under oblique views after opening. Moreover, large 3D asset generation models like TRELLIS provide limited support for interior structures; when doors open, 3D-Gaussians on the inner door panel and the vehicle cavity often appear cluttered. In future work, we plan to incorporate stronger structural priors to better capture the geometric and physical properties of vehicles. One promising direction is to develop regularization strategies that explicitly consider part thickness and enforce surface consistency. For the problem of interior structures, a promising avenue is to leverage diffusion-based inpainting methods to enhance structural plausibility and visual fidelity.

\section{CONCLUSIONS}
In this paper, we present a method for generating faithful and part-controllable 3DGS vehicles pushing the boundary of fidelity and controllability trade-off in car simulation. We integrate TRELLIS asset generation with SAGA segmentation method and 2 PointNet++ models trained on our custom dataset. We also implemented movable part mask insertion and part-edge refinement to enhance the car parts geometry and make them suitable for edition. Our detailed ablation study shows the effectiveness of different components in generating a consistent car asset with the input vehicle after part manipulation of different steering and rotation angles for wheels and front doors. However, we still confirm some space for further improvements in door geometry optimization.


\section*{APPENDIX}

\subsection{PointNet++ Implementation Details}

\subsubsection{Query Labels Prediction}
\label{query_label_prediction_details}
We train a PointNet++ segmentation model with multi-scale grouping (as shown in Fig.~\ref{pointnet2_architecture}) on our custom dataset, using mesh-converted point clouds with ground-truth part labels for point-wise classification. In addition to standard scaling, shifting, and rotation augmentations, each point is randomly perturbed slightly to mitigate the ordered displacement introduced by surface sampling.

Before being input to the model, each vehicle point cloud is downsampled to 8,192 points using Farthest Point Sampling (FPS) during both training and inference. This step reduces the influence of highly dense regions while improving robustness to local displacement patterns. At inference time, the predicted labels for the 8,192 sampled points are propagated to the full point cloud using k-Nearest Neighbors (KNN) \cite{peterson2009k}, enabling reliable part querying of the target car components within the SAGA clusters.

\subsubsection{Kinematic parameters prediction}
\label{kinematic_parameters_prediction_details}
We slightly modified the PointNet++ multi-scale grouping classification architecture by replacing its final MLP classifier with two regression heads: one for hinge axis locations and the other for joint positions. For ease of training, we predict only the 2D projection of hinge axes, since variations along the axis do not affect rotation. In contrast, full 3D joint positions are predicted, as both steering and rotational motions are required for wheels.

\subsection{Experimental Setup}
\label{experimental_details}
Our evaluation uses 7 second-generation vehicle blueprints from CARLA \cite{Dosovitskiy17}: Dodge Charger 2020, Dodge Police Charger 2020, Ford Crown (taxi), Lincoln MKZ 2020, Mercedes Coupe 2021, and Nissan Patrol 2021. Camera viewpoints are defined by 40 poses interpolated between two configurations: \{distance: 5.0, height: 1.1, angle: 270.0, pitch: –10.0\} and \{distance: 5.0, height: 1.5, angle: 270.0, pitch: –20.0\}, with identical positions applied across all pipelines. For part motions, the front left and front right doors are opened, and the front wheels are steered to –1.0, –0.5, 0.5, and 1.0 times their maximum steering angle. In addition, the wheels are rotated by 120° and 240°, resulting in eight distinct poses per vehicle asset.

For evaluation metrics, we use CLIP-ViT-L/14 \cite{radford2021learningtransferablevisualmodels} to compute feature similarity and LPIPS \cite{zhang2018unreasonableeffectivenessdeepfeatures} with an AlexNet backbone for efficient and accurate perceptual distance scoring.


\bibliographystyle{ieeetr}
\bibliography{references}

\end{document}